# Development of a Modular Real-time Shared-control System for a Smart Wheelchair

Vaishanth Ramaraj, Atharva Paralikar, Eung-Joo Lee, Member, IEEE, Syed Muhammad Anwar, and Reza Monfaredi

**Abstract**
In this paper, we propose a modular navigation system that can be mounted on a regular powered wheelchair to assist disabled children and the elderly with autonomous mobility and shared-control features. The lack of independent mobility drastically affects an individual's mental and physical health making them feel less self-reliant, especially children with Cerebral Palsy and limited cognitive skills. To address this problem, we propose a comparatively inexpensive and modular system that uses a stereo camera to perform tasks such as path planning, obstacle avoidance, and collision detection in environments with narrow corridors. We avoid any major changes to the hardware of the wheelchair for an easy installation by replacing wheel encoders with a stereo camera for visual odometry. An open source software package, the Real-Time Appearance Based Mapping package, running on top of the Robot Operating System (ROS) allows us to perform visual SLAM that allows mapping and localizing itself in the environment. The path planning is performed by the move base package provided by ROS, which quickly and efficiently computes the path trajectory for the wheelchair. In this work, we present the design and development of the system along with its significant functionalities. Further, we report experimental results from a Gazebo simulation and real-world scenarios to prove the effectiveness of our proposed system with a compact form factor and a single stereo camera.

## 1. Introduction

The world health organization has estimated that over 65 million people worldwide require a wheelchair for self-mobility [1]. Powered wheelchairs are among the commonly used assistive devices for the personal mobility of physically challenged people. Although current powered wheelchairs do address mobility issues, a majority of disabled population finds them difficult to operate due to physical, perceptual, or cognitive deficits. Several systems have been introduced for wheelchairs to make them autonomous but have their own demerits of being bulky or expensive.

Several studies have shown that an independence in mobility reduces dependence on caregivers and family members, and hence increases vocational and educational opportunities thus promoting a feeling of self-reliance [1, 2, 7]. For young children, independent mobility is vital as, since it forms the foundation for being self-ambulatory [3], whereas for adults it is a matter of self-esteem within the society.

A survey conducted among 200 practicing clinicians [5] indicated that a significant number of disabled individuals have several common difficulties when controlling a wheelchair. Around 9-10% of patients with powered wheelchair training found it extremely difficult to use the wheelchair for their mundane activities. These statistics see a significant jump to 40%, when asked specifically for the steering and maneuverability of the wheelchair. Further, 85% of the clinicians reported patients who lack motor skills, strength, or visual acuity cannot operate a wheelchair.

Nearly half of the patients who are unable to control a powered wheelchair by conventional methods would benefit from an autonomous navigation system [7]. To address these problems several researchers have used existing technologies such as radio frequency identification (RFID)-based tracking, ultrasonic sensors, and gesture control, originally developed for mobile robots to build "smart wheelchairs" [1, 6-8]. In a majority of cases, either a standard powered wheelchair was used as a base on which a system is mounted along with several other sensors, or a seat was attached on top of a mobile robot. A smart wheelchair is a mobile robot and has to be aware of its surroundings using different embedded sensors, local and global path planning, and smart navigation algorithms. It should be able to avoid static and dynamic obstacles and navigate through an environment safely.

Clinicians at the Children's National Medical Center have suggested that total autonomy might not always have constructive effects. It might hinder the growth of motor skills in young children. For instance, cerebral palsy is a group of disorders that affect a person's ability to move and maintain balance and posture. It is the most common motor disability in childhood and accounts for about 1 in 345 children according to estimates from the Center for disease control (CDC) Autism and Developmental Disabilities Monitoring Network [9]. Assistive wheelchair technology must provide a means of mobility for these impaired children without hindering their motor skills growth. Whereas, for elderly people whose developed motor skills are affected due to age might need much more assistance from the navigation system. Hence, to accommodate both age related demographics, a system that can operate both in semi-autonomous and fully-autonomous modes is needed.

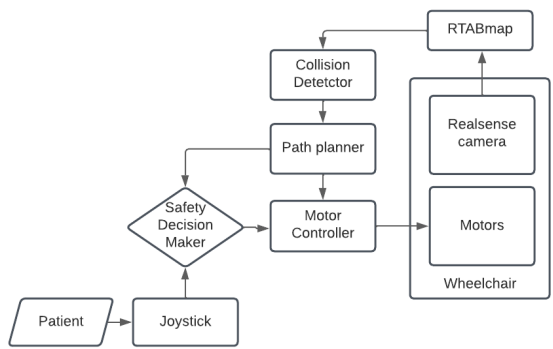

**Fig. 1.** The overall flow of our proposed shared control system.

Hence, in this work, we propose a shared control approach as shown in Figure 1. The system uses a camera and embeds a path planner to share control with the joystick provided with the wheelchair. Particularly, our system is developed to provide a semi-autonomous functionality that can assist a person in navigating through an environment. This is to ensure that children use part of their motor skills to move the wheelchair and at the same time correct their path in case of major deviation from the safe route as shown in Figure 2. Our major contributions are as follows.
- We built a modular system for converting a regular powered wheelchair to a smart wheelchair.
- Our proposed navigation system operates only based on visual odometry.
- We propose a robot operating system (ROS)-based shared control architecture which is essential for children with Cerebral Palsy.

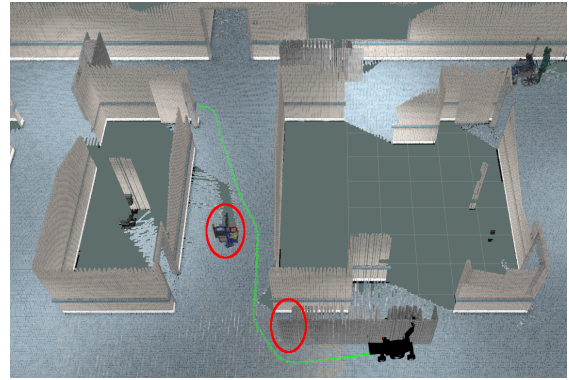

**Fig. 2.** The illustration represents the safely planned trajectory (green) of the wheelchair around the obstacles (red) by our proposed system. In semi-autonomous mode, the user is in control of the wheelchair's motion and any major deviation from the safe path alerts the system to correct to a safe trajectory.

The remainder of the paper is structured as follows. Section 2 discusses related work to create a smart wheelchair with a focus on different populations and applications. Section 3 explains our approach for solving the power wheelchair problems. Section 4 explains the implementation of the smart wheelchair. Section 5 demonstrates the validation of the concept in a simulation environment.

## 2. Related Work

Several researchers have developed various techniques to build an effective and reliable smart wheelchair system to aid people with disabilities. Tomari et al. [6], proposed an intelligent framework to provide a hierarchical semi-autonomous control strategy. Their system provides an interface by detecting head tilt and facial expressions to control the wheelchair's heading. It uses a combination of semi-autonomous and manual approaches to navigate to the desired position. Various sensors were used including a webcam, RGB-depth camera, laser range finder, and an inertial measurement unit (IMU) to perform the semi-autonomous navigation tasks. But since the whole setup consists of several sensors and a bulky computer to process the data it is neither modular nor easy to install.

Simpson et al. [7], proposed a system called "The Hephaestus" which consists of a series of components needed to be added to an existing powered wheelchair to provide assistive navigation features. Their setup consists of 24 sonar sensors for obstacle detection (1-meter range), a bump sensor which is a simple contact switch for detecting collisions, and a laptop that was used to process the data received by the National Instruments data acquisition card from the sensors. At a first glance, this system seems to be robust for navigating a wheelchair but comes with its own flaws. However, multiple blind spots were created due to interference in the signals received by the sonar sensors. The entire processing was done on a laptop which is not only bulky but also consumes a lot of energy.

Kuno et al. [8], implemented a system to detect face and hand gestures to control the wheelchair which is a useful feature for people with severe motor functional problems. A computer with a real-time image processing board was used along with 16 ultrasonic sensors to navigate through the environment including video cameras to detect gestures. To avoid false gesture detection, the video camera feed was used along with ultrasonic sensor data to prevent the wheelchair from turning into an obstacle. Even though the system seems to be useful for patients with severe motor disabilities, it is very specific to that specific category of patients. Detecting face gestures has a lower success rate since human gestures are complex and can be falsely interpreted in multiple scenarios. The whole system with ultrasonic sensors, a video camera, and a personal computer is a very bulky setup.

On comparing our work with the previously mentioned work, our system uses Jetson Nano for performing computer vision tasks which is an inexpensive and low-powered solution. We make use of Realsense depth cameras thereby eliminating the use of bulky and expensive LIDAR or Ultrasonic sensors [8] for navigation and obstacle detection. Since our system aims to be a modular component of a powered wheelchair, it facilitates easy installation for the user. Our system also provides different modes of autonomy which are discussed in section 3.3.

In this work, our proposed system attempts to overcome some of the flaws present in existing systems. Towards this, a vision-based approach is used to perform obstacle detection and path planning. Our design provides shared control and is developed to be used for both pediatric and adult populations.

## 3. Methods

Our proposed system consists of the essential components found in a mobile robot system. In particular, we created a modular design such that the system can be mounted onto a powered wheelchair with little to no modifications. Towards this, we propose to use a vision system and IMU sensors only to localize the robot, rather than using wheel encoders and Lidar sensors. This will allow for an easy integration and a modular system thus streamlining the conversion of a regular powered wheelchair to a smart wheelchair.

The main hardware used is the NVIDIA Jetson Nano(https://developer.nvidia.com/embedded/jetson-nano-developer-kit) which has sufficient processing power (with a graphical processing unit) to enable real-time task planning. To eliminate the bulkiness of the system, various active sensors required for different tasks are replaced with a stereo camera setup, thereby making it compact and inexpensive. Visual odometry plays a very crucial aspect in our system since the mapping and localization are done using visual feedback rather than opting for the bulky and expensive Lidar sensors or wheel encoders.

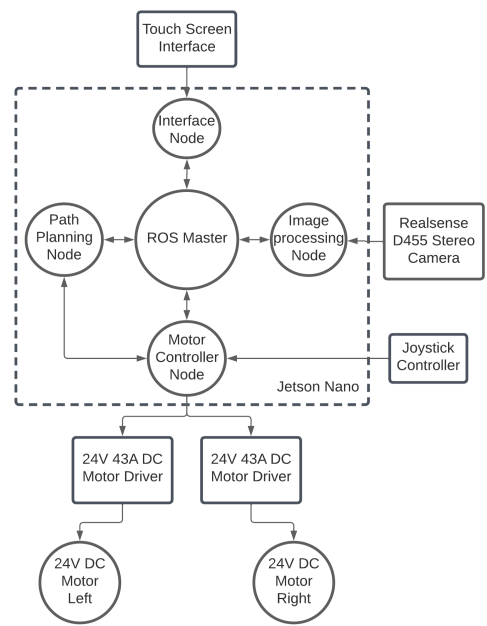

**Fig. 3.** The proposed wheelchair system architecture.

The robot operating system is used as the base platform on which other applications run. ROS is widely used in the robotics community for its flexibility and features. It provides compatibility with various hardware and software hence, providing an easy setup to interface with hardware. The proposed system, as shown in Figure 3, is based on the ROS architecture with four nodes including a) Interface system node, b) Image processing node, c) Path planning node, and d) Motor controller system node. Each of these nodes serves a specific purpose, and some nodes rely on other nodes to conduct their tasks. The specific functionalities of each node are presented in the following subsections.

### 3.1 Interface System

This node acts as a bridge between other nodes to coordinate their tasks and provide a collective output. It is also responsible for taking user inputs according to which the system acts.

A kivy-python-based package has been developed to provide the user interface for controlling manual, semi-autonomous, and autonomous modes. The different modes of operation are discussed in section 3.3. The interface was implemented and tested in simulation. The package is cross-platform and capable of taking touch inputs from the user. The final prototype will have a touch-based LCD screen to control the smart wheelchair. The display will show a 2D map of the environment that's mapped by the system, which can be utilized by the user to select checkpoints where the user intends to go.

### 3.2 Image Processing

The crucial aspect of reliably navigating a system through an environment is to map and localize the object within the environment. This might be a challenging task since the environment in a real-world scenario is dynamic and can cause map distortions.

It is common to use a LIDAR sensor to map an environment, but it is a bulky and expensive solution. Hence, using a stereo camera to map an environment in real-time would be effective. The concept of visual odometry is used wherein the depth data from the stereo camera and the robot orientation from the IMU sensor are used in conjunction to create a 3D map of the environment. Various open-source simultaneous localization and mapping (SLAM) applications are available, but two essential requirements must be considered to make a selection. First, the SLAM algorithm needs to be well-performing, actively maintained, and needs to support the chosen algorithm and ROS. Secondly, the sensor needs to be supported by the library, should be easy to configure and set up, and has a good price-performance metric. Real-Time Appearance Based Mapping (RTAB-map) is chosen since it fulfills the above-mentioned criteria and is widely used for visual odometry. The RTABmap creates and stores the mapped environment in a database of images with the camera orientation and matching key points. Using RTABmap, the localization could be performed by only using visual data and IMU sensor data eliminating the need for a wheel odometer, thus making the system installation easy.

Once the environment is mapped, the RTABmap can localize the robot using the camera feed by performing image matching with the existing database and also can provide the orientation of the robot. For the robot to move safely in a dynamic environment, it should be aware of its nearby obstacles. RTABmap provides an application programming interface (API) for obstacle point clouds which are processed using the depth image data from the stereo camera. The obstacle point cloud, as shown by the white dots in Figure 4, can be used to detect and categorize the obstacles, thereby providing a safer way to navigate the system around the obstacle.

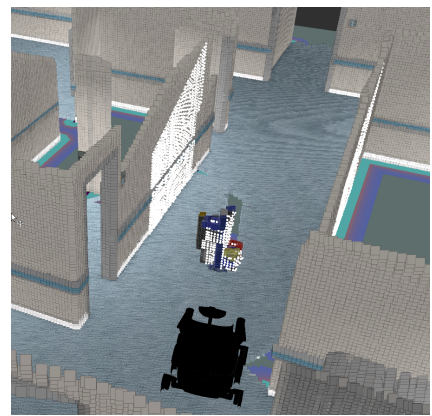

**Fig. 4.** The projected white point cloud represents the obstacle point data that is used for collision detection and obstacle avoidance.

### 3.3 Path Planning

The user is provided with manual, semi-autonomous, and autonomous control modes. The manual mode allows the user to control the wheelchair by providing directional commands from the interface. In semi-autonomous mode, the user selects a destination and drives to it. The system creates a safe path connecting the destination and the current wheelchair position but interferes with the control only if the user diverges from the path beyond the threshold limit. This is feasible for disabled children who are required to train their motor-muscle skills which are vital for the child's growth. Whereas in the autonomousmode, the system takes over the entire control and moves the wheelchair to the user-defined destination on the 2D map.

The virtual map created by the RTABmap provides a solid base on which we can use various path planning algorithms to create a feasible path from the current location to the desired destination. ROS comes bundled with a package called move base which can be used to calculate a safe and quick path.

When calculating a path, there are physical system constraints to be considered to prevent collision with walls and other static objects. The move base package creates a costmap by taking in sensor data from the image sensor that enables it to take the physical constraints of the system into account. A cost map as shown in Figure 5 is basically a 2D occupancy grid of the data. A 2D occupancy grid is a 2D representation of the map where each pixel denotes three of the following states, free, occupied, or unknown. It basically determines if it's possible for the wheelchair to move in a specific area. A 2D costmap is constructed based on the occupancy grid which incorporates the inflation layer. The inflation layer can be understood as padding applied to the walls and corners to prevent the robot from colliding or coming in close contact with them. The inflammation layer is determined by the inflation radius provided by the user or the application developer. The inflation radius is always greater than the width of the wheelchair. It is always a good practice to have an inflation radius that is 1.2 times the width of the wheelchair. Thus ensuring the optimal costmap decay curve is one that has a relatively low slope as mentioned by Dr. Pronobis in his move base navigation guide. [12]

While each cell in the costmap can have one of 255 different cost values, the underlying structure that it uses can represent only three states. It can be either free, occupied, or unknown. Once the user has selected a destination on the 2D map the move base algorithm efficiently plans a path as shown in Figure 6. Hence, the move base package is able to navigate safely in the environment on the user's command.

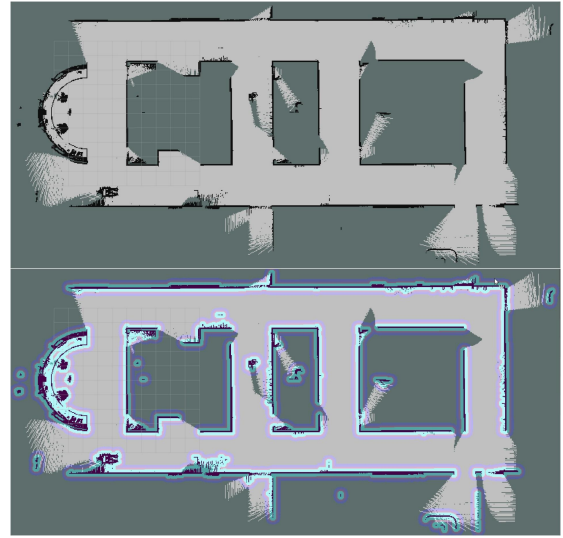

**Fig. 5.** Side-by-side comparison of the original map and the costmap. The blue layer represents the inflammation layer, the walls are represented by black lines and the gray area indicates the safe area to plan paths.

The move-base package provides an inbuilt recovery behavior for situations where the wheelchair is stuck in an obstacle zone. By default, the move_base node will take the following actions to attempt to clear out space. First, obstacles outside of a user-specified region will be cleared from the robot's map. Next, if possible, the robot will perform an in-place rotation to clear out space. If this too fails, the robot will more aggressively clear its map, removing all obstacles outside of the rectangular region in which it can rotate in place. This will be followed by another in-place rotation. If all this fails, the robot will consider its goal infeasible and notify the user that it has aborted. These recovery behaviors can be configured using the recovery_behaviors parameter and disabled using the recovery_behavior_enabled parameter. [13]. But sometimes the algorithm still struggles to get the wheelchair out of such scenarios. In that case, our algorithm detects that the wheelchair has not performed any motion for a threshold time period and resets the move base algorithm, and plans another path.

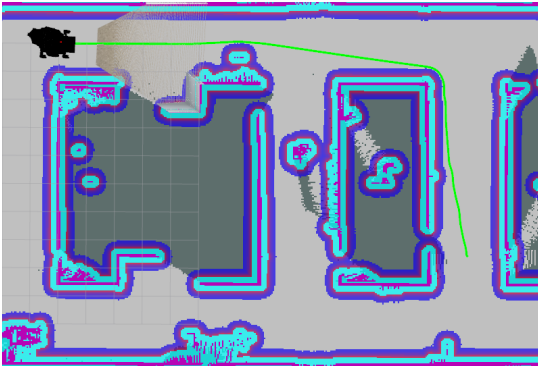

**Fig. 6.** The planned path for the given destination point. The blue layer surrounding the walls represents the inflation layer in the cost map to prevent the wheelchair from moving too close to the walls or other obstacles.

One of the inconveniences that come with using ROS1 for any system has to be looked at is the pseudo-real-time architecture of ROS1 itself. The absence of a task scheduler and the fact that it is built on top of Linux makes ROS1 less equipped for strict time-bound requirements. While ROS is fast and used for several robot operations, a small control cycle delay exists in every ROS application. However, during rigorous testing in our simulations, it was observed that this tiny delay is negligible since the real-time factor in Gazebo never dropped below 1, which represents a real-time operation. Based on this observation, we argue that our proposed system will not have any significant issues while performing its operation in real-time.

The ability of the package within ROS (move_base) to plan in real-time is dependent on the underlying hardware. However, for our application the smart wheelchair moves at relatively slower speeds (simulation velocity of 0.5 m/s) thus a limited latency could be acceptable.

### 3.4 Motor Controller System

This node is responsible for providing pulse width modulated signals to the motor driver that in turn actuates the motor at different speeds. The system's motor is based on simple differential drive actuation. The input from the joystick controller is also processed by this node for manual control over the system. For disabled children, it is recommended to drive the wheelchair on their own to train their motor muscle skills. But in situations where the user losses control over the wheelchair's motion in case of accidental control input, the semi-autonomous mode would take control of the wheelchair and correct its heading based on the safe path computed previously. The autonomous mode is particularly aimed at disabled elderly people who have a tough time controlling the wheelchair. Hence, they can input their destination and let the autonomous system drive the user.

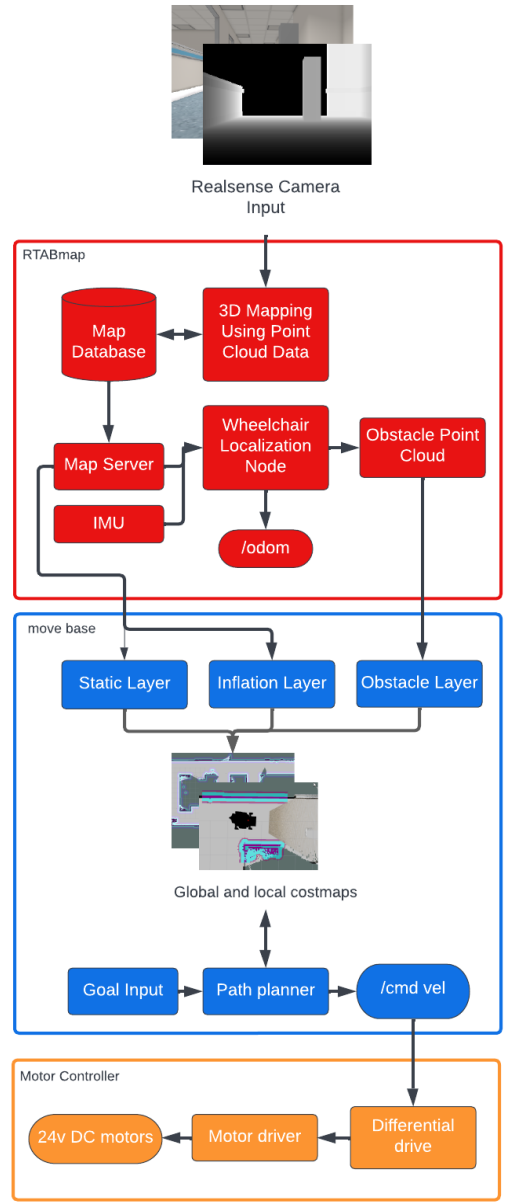

**Fig 7**. The overall workflow of the system.

The controller receives the command input from the safety decision-maker system as shown in Figure 1. The safety decision maker feeds either output from the joystick device or the path planner node based on the collision prediction and assigned safety margin.

## 4. Hardware and Software Implementation
### 4.1 Hardware
In our work, the system is intended to be modular such that to provide for an easy installation on an already existing powered wheelchair. The image processing computation is carried out on the Jetson Nano hardware which provides the much-required GPU computational power required by the system. To keep the form factor small and provide sufficient computational power, our system consists of a Jetson Nano 2GB developer board at its heart. The Jetson Nano is 2.72 x 1.77 x 1.77 inches in dimension and weighs less than 100g making the system compact and efficient.

It is powered by a compact 8 volt (V) power module attached to the Nano's base. Nvidia's Jetson platform can handle computer vision and path planning tasks in real-time with minimal lag. The setup is equipped with an Intel RealSense stereo camera module that provides with vision data required to map the environment. Two separate 43 Ampere 24V DC motor drivers are used to control the right and left 24V DC motor. These components are mounted on a powered wheelchair.

### 4.2 Software
During the first startup, the robot is completely blind since it has no map to localize itself. Hence, it is changed to mapping mode wherein the robot moves around to map the environment. Even when there is no map to give an estimate of the nearby obstacle the RTAB-map provides an API for obstacle point cloud using which we can map the environment safely. When the environment is sufficiently mapped it takes approximately the same path back to where it started to perform loop closure. This allows tuning out the irregularities in the map due to a drift in the IMU sensor data.

Once done, the user can set it to the localization mode after which the wheelchair can localize in the environment using visual odometry.

## 5. Experiments
### 5.1 Simulations
Simulators are very powerful tools to test out prototypes in various real-world scenarios which might be difficult to physically recreate due to cost and availability concerns. ROS provides a simulation package called Gazebo that can simulate various environments in real time[11].

The wheelchair system was simulated in a hospital environment and a home-like environment obtained from AWS Robotics' open-source github repository as shown in Fig 8 to test out the capabilities. The system was able to map and localize in its environment as well as to detect nearby obstacles and avoid collisions with them.

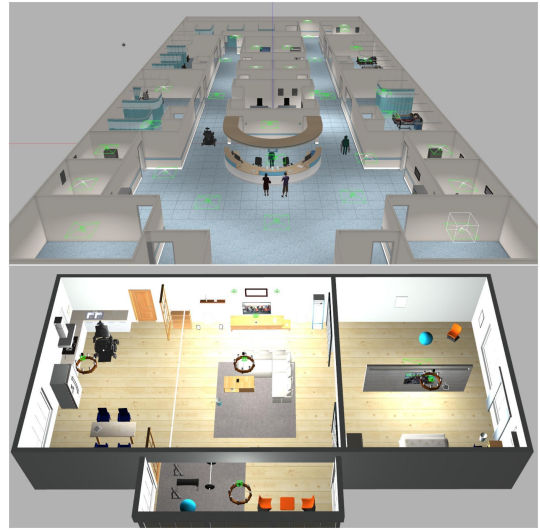

**Fig. 8.** The custom world imported into Gazebo. Green boxes in the first image indicate the ceiling lights.

A realistic model of the wheelchair was modeled and imported into the Gazebo simulation environment as shown in Figure 9. It is driven by a differential motor driver provided by the ROS package. An RGBD camera plugin is mounted in front of the wheelchair as highlighted in Figure 9, and it is aligned so that it would not obstruct the user's view.

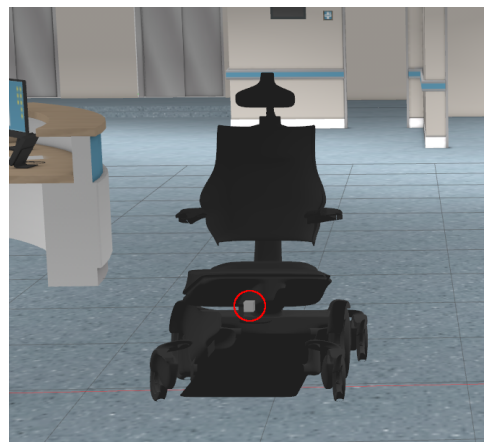

**Fig. 9.** A detailed model of the wheelchair. The red circle highlights the RGBD sensor mounted on the wheelchair.

It is crucial to test the controller's ability to navigate through dynamic obstacles. The gazebo provides a plugin to create dynamic obstacles and provide a trajectory to it. The dynamic obstacles were placed in the environments as shown in Figure 10, such that it mimics the behavior of a human being doing mundane things.

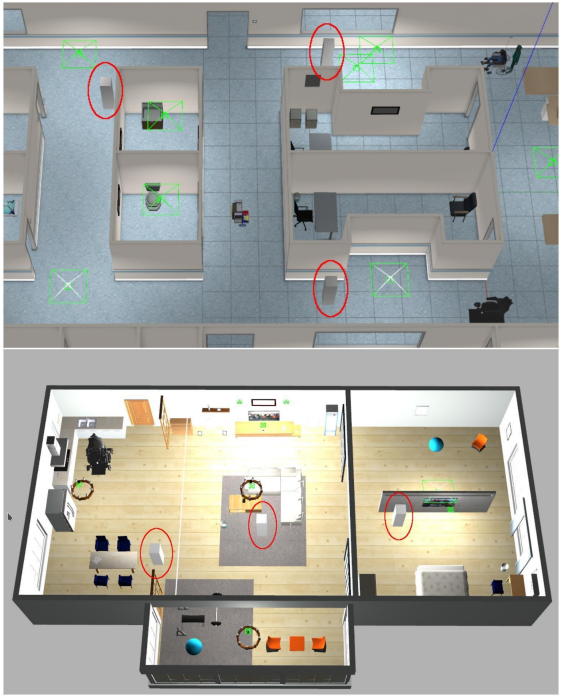

**Fig. 10.** Both images denote the presence of dynamic obstacles in bost hospital and home-like environments. The red circle highlights the dynamic obstacles.

The path planning performed by the move base package can be visualized in RVIZ. The path planning node takes care of the integration of user input to the move base package. Both the local and global costmaps are updated to accommodate for the changes in the feasible path planning area. The obstacle cloud points provided by the RTABmap API are processed to update the costmaps. The costmaps are updated when there is an obstacle in the threshold value which represents the distance of the obstacle from the wheelchair. It is also able to clear areas of the cost map that previously had an obstacle but are free now.

### 5.1.1 Quantitative measures

To evaluate the effectiveness of the wheelchair's control system, simulation trails were performed with both specific goal points and random goal points. The robot moved with a maximum linear velocity of 0.5m/s and maximum angular velocity of 1.5 m/s. The start and end locations are selected manually such that it takes the following criteria under consideration. Firstly, it must not be a simple straight path but must have turns around the corners and should pass through a narrow region. secondly, It must have either a static or dynamic obstacle along the path. Thus we selected two paths considering the above criteria and ran trials on that paths. The measured results are shown below.

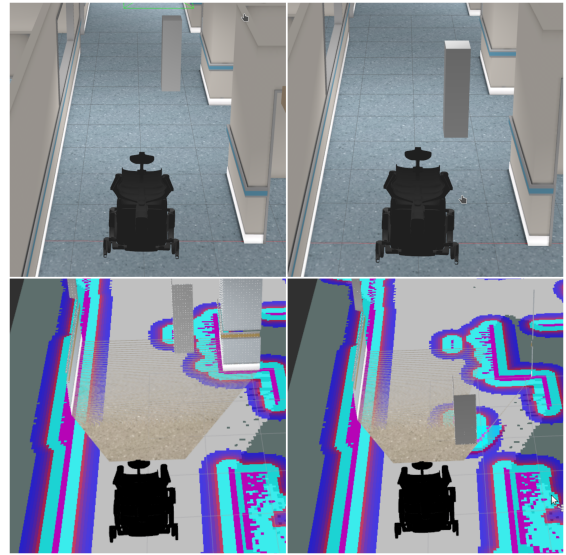

**Fig. 11.** The costmap being updated based on the nearby dynamic obstacle using the cloud point data.

**Table 1.** Simulation trials involving the wheelchair to plan and navigate to goal location 1 with static obstacles.

| Trial No. | Distance (m) | Real Time Taken (secs) | Goal reached | Remarks |
|---|---|---|---|---|
| 1 | 28.40 | 150 | True | - |
| 2 | 27.50 | 155 | True | Spinned 2 times near the wall |
| 3 | 28.50 | 146 | True | - |

It took an average time of 150.3 seconds to reach the goal at an average distance of 28.13 meters.

**Table 2.** Simulation trials involving the wheelchair to plan and navigate to goal location 2 with static obstacles.

| Trial No. | Distance (m) | Real Time Taken (secs) | Goal reached | Remarks |
|---|---|---|---|---|
| 1 | 23.85 | 135 | True | - |
| 2 | 24.20 | 133 | True | - |

| 3 | 23.50 | 140 | True | - |

It took an average time of 136 seconds to reach the goal at an average distance of 23.85 meters.

**Table 3.** Simulation trials involving the wheelchair to plan and navigate to random goal points with static obstacles.

| Trial No. | Distance (m) | Real Time Taken (secs) | Goal reached | Remarks |
|---|---|---|---|---|
| 1 | 23.41 | 189 | True | - |
| 2 | 15.39 | 202 | True | Struggled to avoid colliding with the walls |
| 3 | 17.41 | 160 | True | - |
| 4 | 25.23 | 192 | True | - |

**Table 4.** Simulation trials involving the wheelchair to plan and navigate to goal location 1 with dynamic obstacles.

| Trial No. | Distance (m) | Real Time Taken (secs) | Goal reached | Remarks |
|---|---|---|---|---|
| 1 | 29.20 | 170 | True | - |
| 2 | 28.50 | 165 | True | - |
| 3 | 28.40 | 250 | False | Collided with obstacle and got stuck |

Out of the two successful attempts it took an average time of 170 seconds to reach the goal at an average distance of 28.85 meters.

**Table 5.** Simulation trials involving the wheelchair to plan and navigate to goal location 2 with dynamic obstacles.

| Trial No. | Distance (m) | Real Time Taken (secs) | Goal reached | Remarks |
|---|---|---|---|---|
| 1 | 24.50 | 150 | True | - |
| 2 | 24.20 | 300 | False | Collided with second obstacle |
| 3 | 23.80 | 350 | False | Collided with first obstacle and the walls |

In this trail it was able to achieve success in only one trial because of the complexity of the path. Since the path had two dynamic obstacles moving in different directions, the wheelchair was able to avoid one obstacle but collided with the other.

**Table 6.** Simulation trials involving the wheelchair to plan and navigate to random goal points with Dynamic obstacles.

| Trial No. | Distance (m) | Real Time Taken (secs) | Goal reached | Remarks |
|---|---|---|---|---|
| 1 | 14.39 | 250 | True | - |
| 2 | 24.42 | 210 | True | Moved very close to obstacle |
| 3 | 18.52 | 400 | False | Failed to reach the goal because of collision |
| 4 | 26.31 | 255 | True | Got stuck near the wall for few seconds |

To summarize, the wheelchair achieved a 100% success rate while reaching a goal in a static environment. It did struggle sometimes to align itself in the direction of the path but it managed to recover from the situation and reach the goal. But in the case of dynamic environments, the wheelchair was only able to achieve a 70% success rate. This proves that the wheelchair control algorithm is able to detect the dynamic obstacle and navigate around it but still requires fine tuning to improve the success rate.

### 5.2 Hardware Implementation
#### 5.2.1 Hardware Setup
Testing out the system in the real world was a challenging task. Because of the high risk of damaging the powered wheelchair during the testing process, we decided to build a basic model as shown in Figure 12 that represents the base of the wheelchair. The motors and the battery setup resemble that of the powered wheelchair which is rated at 24V DC. A wooden plank is mounted in the front of the battery to hold the system modules.

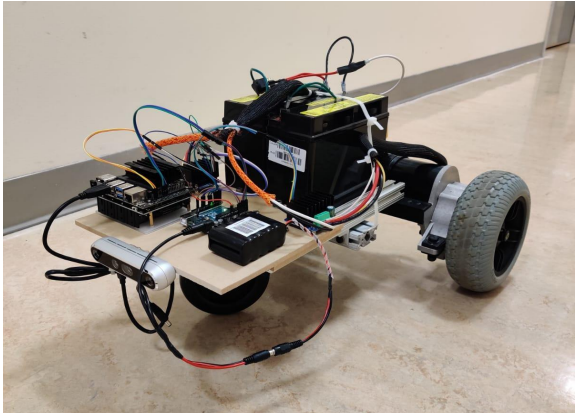

**Fig 12.** The basic hardware prototype with the system modules mounted on the front and a 24V DC motor setup at the back.

The hardware setup consists of a 24V DC motor powered by 2 series connected 12 V battery with max current input of 18 A as highlighted in red and blue in Figure 13. The motors are equipped with electronic brakes and a lever mechanism to manually actuate brakes. The brakes can be disengaged by providing 24V DC power to the brake terminals. A 24V 30A PWM motor driver is used to actuate the motor drivers based on the control input from Jetson Nano.

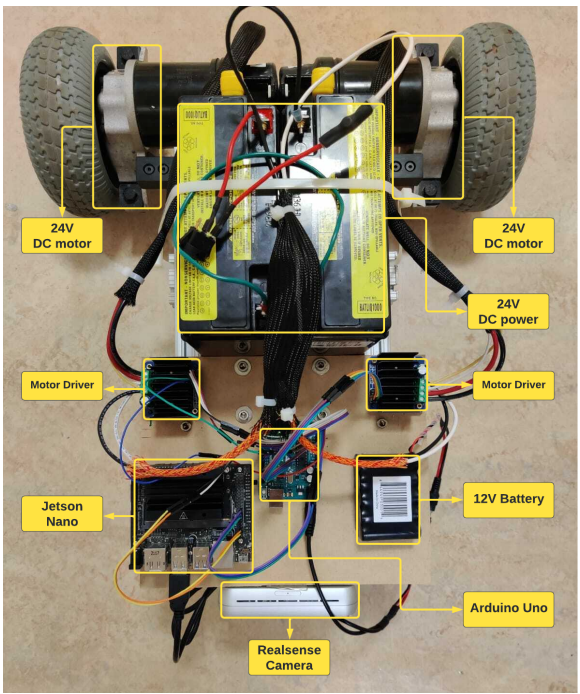

**Fig. 13.** An image labeling the different components of the hardware system.

Arduino Uno is used to deliver the signal to the motor driver. A 12v DC power supply is used to supply power to the Arduino Uno and Jetson Nano has a separate 8V power module mounted at its bottom to provide a compact form factor. The Realsense camera is mounted on the front making sure it has the best visibility of the environment. The stereo camera data is relayed to the Jetson Nano using a USB-C cable.

### 5.2.2 Real-Time Mapping and Localization

The mapping and the localization were completely performed on the Jetson nano. We were able to map our lab and navigate around it. RTABmap processes the RGB and the depth images captured by the Realsense camera. Features are extracted from the captured images using the GFTT (Good Features to Track) and ORB (Oriented FAST and Rotated BRIEF) algorithms as shown in Figure 14. The extracted features enable the algorithm to compare it with consecutive images that are in turn used for localizing the robot as shown in Figure 15. The images are captured at a frequency of 5Hz and are stored in the database along with the camera's intrinsic and extrinsic factors.

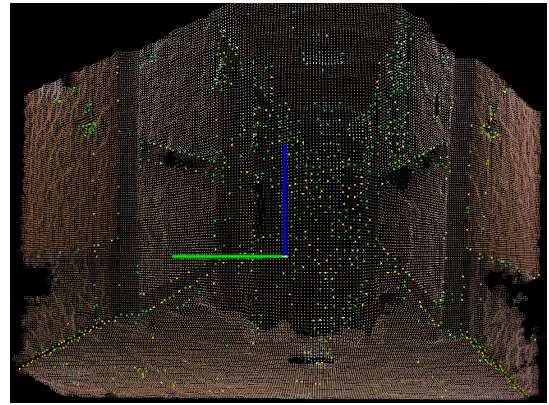

**Fig. 14.** The features extracted from the input RGB-D camera data.

In very up close and complicated situations the algorithm finds it difficult to localize itself due to a lack of features from the RGBD images. But the algorithm manages to solve the issue by making use of data from an IMU sensor embedded in the Realsense camera. In case both situations fail the algorithm updates its location when the robot moves further and detects enough features on the RGBD image. The green and red axis, shown in Figure 14, represents the robot`s pose and orientation.

In a real-world scenario, there are a lot of constantly moving obstacles. It is necessary to map the environment such that it removes moving objects from the scenario and takes only the static objects into the algorithm, thus making it easy to use the final 2D map to plan the path. The algorithm performs very well in such scenarios and was able to remove the dynamic obstacle from the mapped floor plan.

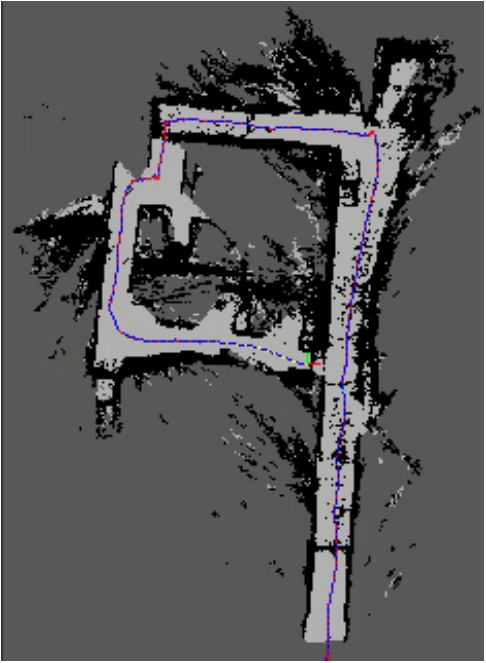

**Fig. 15.** The wheelchair localizes itself in the mapped environment using the stereo camera data also taking its orientation into account.

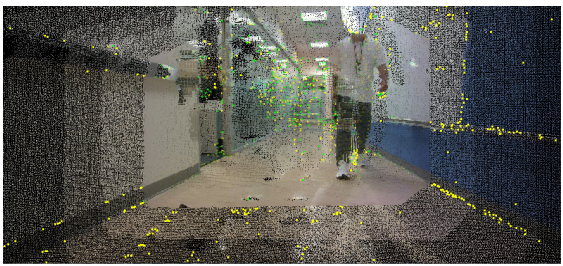

**Fig. 16.** A scenario wherein a person walks into the frame while mapping the environment. The RTABmap algorithm manages to remove the dynamic obstacle from the map.

## 6. Limitations

Our objective was to design and develop a comparatively inexpensive and compact system to mount on a powered wheelchair. The approach we took was to minimize the number of sensors and use a sufficiently powerful yet compact microcomputer. The Realsense camera is our primary sensor. Our system comes with the following limitations.

1. Since only one stereo camera sensor is mounted in the front of the wheelchair, any obstacle not visible in the camera frame cannot be processed as a dynamic obstacle unless it comes inside the frame.
2. Relying solely on the RGBD data instead of a physical odometer can sometimes be unreliable and the wheelchair might find it difficult to localize itself.
3. RGBD cameras provide depth up to a certain distance of 3 m in the case of Intel RealSense, depth estimate is often noisy and has a limited field of view.
4. Even Though a move base can quickly and safely plan a path, it lacks the ability to navigate the robot in narrow obstacle space and often get stuck. The algorithm's recovery behavior mentioned in section 3.3 kicks in but sometimes the algorithm still struggles to get the wheelchair out of such scenarios.

## 7. Conclusion

In this paper, we introduce a modular system that can be mounted on an existing powered wheelchair to provide autonomous assistive functionality to the disabled user. The system is designed such that it is compact and cost-efficient but at the same time delivers the required support for the users. Making full use of machine vision is difficult, but it can provide various opportunities as well as cut down the use of bulky and expensive sensors. By using open-source software, we provide developers to create their own applications to improve the product. Our wheelchair system is intended to be used in a dynamic environment, which may be a hospital or an indoor space inside a house. It will be able to assist the user to navigate to their specified destination with ease and safety.

## 8. Technical Biography

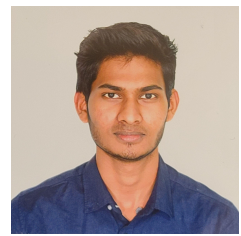

Vaishanth Ramaraj is a graduate student pursuing a Master's in Robotics Engineering from the University of Maryland, College Park, MD, USA. Currently, he is an intern at Children's National Medical Center, Washington DC. He is an avid robotics application developer and worked on several

ROS-related projects. His main focus is on developing computer vision and path planning applications. He published a Gedrag & Organisatie journal paper on "Design and Development of Spherical Robot using Pendulum Mechanism" in 2020.

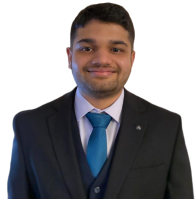

Atharva Paralikar is a graduate student pursuing a Master's in Robotics Engineering at the University of Maryland, College Park, MD, USA. He is an enthusiastic roboticist and has worked on various ROS-based projects. Previously he has also worked on designing robots for Automotive Spray Painting applications. His current focus is on developing medical and wearable robotics.

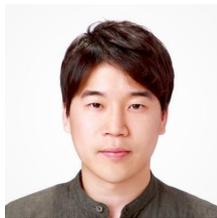

Eung-Joo Lee received his Ph.D. degree from the University of Maryland, College Park, MD, USA in 2021. He performs research on computer-aided design and implementation of a digital signal processing system. Specifically, he focuses on real-time computer vision and medical imaging applications under challenging constraints based on deep learning techniques. He is currently a postdoctoral research fellow at Massachusetts General Hospital/Harvard Medical School.

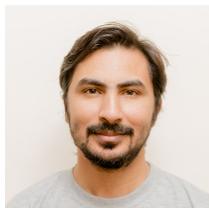

Syed Muhammad Anwar received a B.Sc. degree (Hons.) in computer engineering from the University of Engineering and Technology (UET), Taxila, Pakistan, M.Sc. degree (Hons.) in data communications and the Ph.D. degree in electronic and electrical engineering from The University of Sheffield, U.K., in 2012. He is currently an Associate Professor at Children's National Hospital, Washington DC. His research interests include medical imaging, data communication, machine learning, and human–computer interaction.

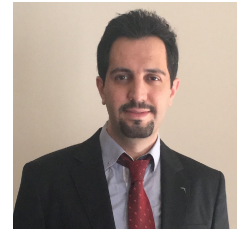

Reza Monfaredi received his Ph.D. degree from Amirkabir University of Technology, Tehran, Iran in 2011. Currently, he is a faculty member at Children's National Medical Center, Washington DC where he completed his postdoctoral fellowship in 2013. He also is an adjunct assistant professor at Maryland university college park. His main focus now is MRI-compatible robotics, rehabilitation robotics, and Medical devices.

## 8. Statements and Declarations
### 8.1 Competing Interests
All authors declare that they have no conflicts of interest that are relevant to the content of this article.

Funding: This project was internally funded by Sheikh Zayed Institute for Pediatric Surgical Innovations at Children's National Medical Center.

### 8.2 Author contributions
Vaishanth Ramaraj implemented the ROS based control system. Vaishanth Ramaraj and Atharva Chandrashekhar Paralikar contributed to the system integration and the hardware implementation. Eung-Joo Lee provided technical supports on the real-time aspects of the project. Syed Muhammad Anwar provided technical supports on image processing aspect of the project and Reza Monfaredi provided technical supports on system integration and shared control system.

### 8.3 Data Availability
The datasets generated during and/or analyzed during the current study are available from the corresponding author on reasonable request.

### 8.4 Ethics Approval

Not applicable